\documentclass[conference]{IEEEtran}
\IEEEoverridecommandlockouts
\usepackage{cite}
\usepackage{amsmath,amssymb,amsfonts}
\usepackage{algorithmic}
\usepackage{graphicx}
\usepackage{textcomp}
\usepackage{xcolor}
\usepackage{tikz}
\usetikzlibrary{arrows.meta}
\usepackage{subfig}
\usepackage{float}
\usepackage{placeins}
\usepackage{stfloats}
\usepackage[T1]{fontenc}

\newcommand{\orcidlink}[1]{}

\def\BibTeX{{\rm B\kern-.05em{\sc i\kern-.025em b}\kern-.08em
    T\kern-.1667em\lower.7ex\hbox{E}\kern-.125emX}}
\begin{document}

\title{Distributed Coordination for Resilient Multi-UAV Remote Sensing: A Photovoltaic Inspection Case Study%
}


\author{
\IEEEauthorblockN{                                  
Guillermo GP-Lenza\IEEEauthorrefmark{1}\IEEEauthorrefmark{2}, Miguel Fernandez-Cortizas\IEEEauthorrefmark{1}\IEEEauthorrefmark{3}\orcidlink{0000-0002-3822-075X},
Martin Molina\IEEEauthorrefmark{2}\orcidlink{0000-0001-7145-1974},
Pascual Campoy\IEEEauthorrefmark{1}\orcidlink{0000-0002-9894-2009}}
\IEEEauthorblockA{\IEEEauthorrefmark{1}\textit{Centre for Automation and Robotics C.A.R, (UPM-CSIC), Universidad Polit\'{e}cnica de Madrid,} Spain}              
\IEEEauthorblockA{\IEEEauthorrefmark{2}\textit{Department of Artificial Intelligence, Universidad Polit\'{e}cnica de Madrid,}, Spain}              \IEEEauthorblockA{\IEEEauthorrefmark{3}\textit{Automation and Robotics Group (ARG-SnT), University of Luxembourg - SnT,} Luxembourg}} 

\maketitle

\begin{abstract}
Deploying multiple UAVs for remote sensing enables proportional reductions in mission time, but realizing these benefits requires the fleet to coordinate at runtime: distributing sensing targets, responding to platform failures, and recovering from degraded data quality. In inspection campaigns, where mission value depends on complete coverage and the usability of every capture, a centralized ground-station coordinator is a single point of failure: a lost link or station fault leaves sensing gaps that cannot be filled without operator intervention. We propose the \textbf{SwarmLink}, an inter-agent communication infrastructure that non-invasively extends any existing aerial framework with peer-to-peer coordination capability, without modifying the host system. We apply it to photovoltaic plant inspection as a representative large-scale sensing campaign, extending Aerostack2 with a distributed auction that unifies initial sensing-target allocation, platform-failure recovery, and data-quality-triggered reassignment into a single runtime mechanism. All three disruption scenarios reduce to the same re-auction over remaining targets and active platforms, requiring zero modifications to the Aerostack2 core and no ground-station involvement during the mission.
\end{abstract}

\begin{IEEEkeywords}
multi-UAV systems, photovoltaic inspection, distributed task assignment, auction algorithms, inter-agent communication, aerial robotics frameworks, remote sensing
\end{IEEEkeywords}

\section{Introduction}
\label{sec:intro}

Deploying multiple Remotely Piloted Aircraft Systems (RPAS) for remote sensing and inspection offers proportional reductions in mission time and enables coverage of spatially extensive targets. Photovoltaic (PV) plant inspection is a representative example: fleets of UAVs equipped with RGB and thermal cameras cover large panel installations at a fraction of the cost and risk of manual methods \cite{8682048,MICHAIL2024e23983,JaviCPP}. Realizing these benefits, however, requires the fleet to coordinate at runtime, distributing sensing targets, reacting to platform failures, and responding to degraded data quality rather than executing fixed, pre-assigned plans.

Multi-UAV inspection campaigns impose three concurrent requirements: complete coverage of all sensing targets, usable data quality for every acquired capture, and resilience to in-mission platform failures. Current practice addresses none of these simultaneously: tasks are partitioned by a ground-station planner prior to takeoff and distributed as fixed routes \cite{Luna2024,Stodola2025}, with no mechanism for runtime renegotiation when either of the following disruptions occurs.

\begin{figure}[htbp]
\centerline{\includegraphics[width=\columnwidth]{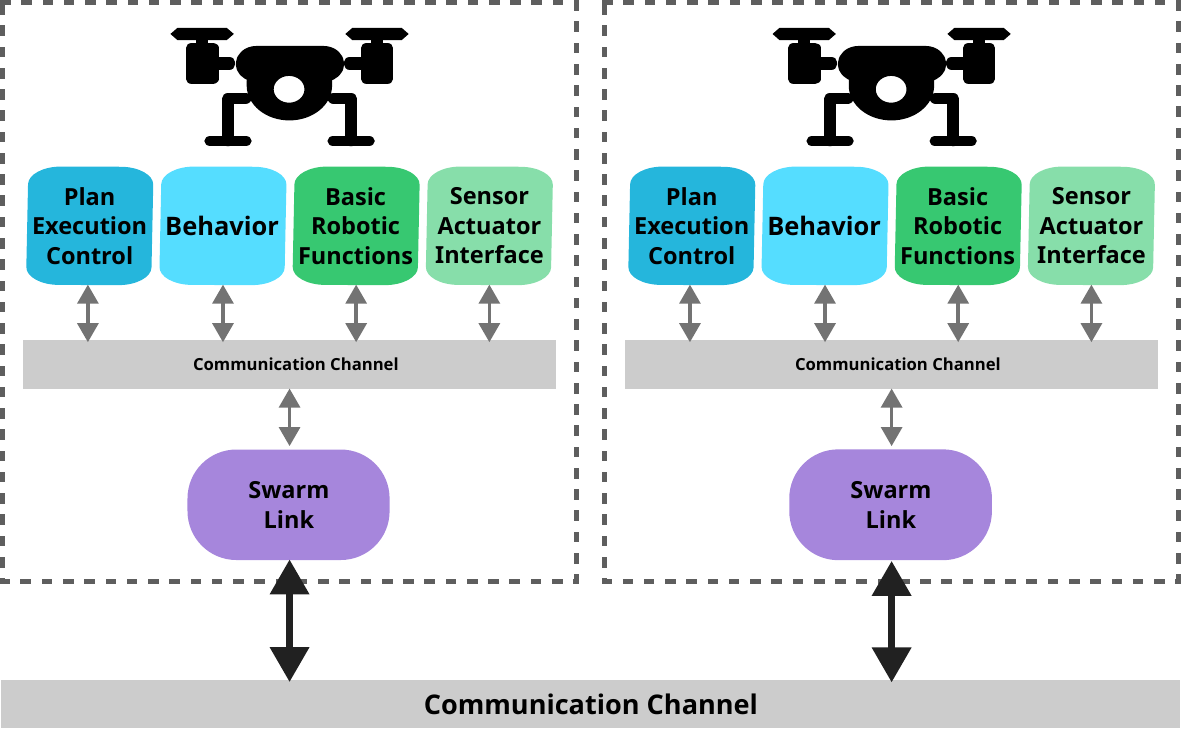}}
\caption{Architecture of a representative aerial inspection framework. Each drone instance is namespace-isolated. The SwarmLink (Section~\ref{sec:gateway}) adds a peer-to-peer coordination layer; the Auction Behavior (Section~\ref{sec:auction}) uses it to allocate and reassign sensing targets across the fleet. \vspace{-0.95em}}
\label{fig:aerostack2}
\end{figure}

\textbf{Platform failure.} UAVs are subject to in-mission failures including battery depletion, GNSS loss, and mechanical faults \cite{Yang2022}. Because assignments are computed and stored centrally prior to takeoff, no surviving agent retains knowledge of the failed drone's pending targets. Recovery requires the ground station to detect the failure, retrieve the abandoned assignments, recompute a partition for the reduced fleet, and push updated plans, introducing latency proportional to ground-link availability and operator response time, with multiple points of failure in the recovery chain.

\textbf{Data quality degradation.} Onboard image evaluation can identify captures degraded by motion blur, adverse illumination, or specular reflections \cite{JaviCPP}. Under current practice, the affected drone logs the event and continues along its pre-assigned route; reassignment is deferred to a post-flight correction cycle that may necessitate a full re-survey of the affected area, with coverage gaps persisting for the remainder of the mission.

Both failure modes require the active fleet to renegotiate target assignments at runtime without ground-station mediation. Current aerial inspection frameworks do not provide this. Systems such as Aerostack2 \cite{fernandezcortizas2024aerostack2softwareframeworkdeveloping}, which organize drone software into abstraction layers (sensor-actuator interfacing, robotic functions, high-level behaviors, plan execution, mission control), run as independent, namespace-isolated instances per drone. A module on one drone has no mechanism to exchange messages with a peer behavior without manual, per-protocol wiring of communication endpoints. Any coordination logic is therefore forced into an external ground station, reintroducing the centralization the distributed approach aims to eliminate.

We address this at two levels through an integrated coordination architecture layered non-invasively on top of the existing framework. At the infrastructure level, we propose the \textbf{SwarmLink}: a framework-agnostic communication layer that gives any behavior a uniform interface, \textit{register} for message types, \textit{forward} messages to peers, without knowledge of the underlying transport or fleet composition. At the coordination level, we apply the SwarmLink to PV inspection in Aerostack2 \cite{fernandezcortizas2024aerostack2softwareframeworkdeveloping} as shown in (Fig.~\ref{fig:aerostack2}), implementing a distributed auction that unifies initial sensing-target allocation, platform-failure recovery, and data-quality-triggered reassignment into a single runtime mechanism. The key architectural insight is that all disruption events reduce to the same re-auction over (remaining targets, active drones); the SwarmLink is what makes this unified, ground-station-free response possible without modifying the host framework. Table~\ref{tab:comparison} positions this work against representative existing systems.

\begin{table}[t]
\caption{Comparison with representative multi-UAV PV inspection systems. GS\,=\,ground station; AS2\,=\,Aerostack2.}
\label{tab:comparison}
\begin{center}
\small
\renewcommand{\arraystretch}{1.2}
\begin{tabular}{p{2.0cm}p{1.55cm}p{1.55cm}p{1.55cm}}
\hline
\textbf{Property} & \textbf{Luna et.al~\cite{Luna2024}} & \textbf{Melero-Deza et.al~\cite{JaviCPP}} & \textbf{Ours} \\
\hline
Task assignment    & Centralized replan      & GS dispatch       & Distributed auction \\
Failure recovery   & Centralized replan      & Not addressed     & Distributed re-auction \\
Data quality       & Not addressed           & Not addressed     & Runtime reassignment \\
GS independence    & No                      & No                & Yes \\
Integration        & Standalone system       & Centralized add-on to AS2 & Non-invasive ext.\ of AS2 \\
\hline
\end{tabular}
\end{center}
\end{table}

\section{SwarmLink}
\label{sec:gateway}

\subsection{Design Concept}

The first contribution of this paper is the \textit{SwarmLink}: an inter-agent communication infrastructure designed around two principles.

The first is \textbf{separation of concerns between use and transport}. Any component of an aerial system that wishes to participate in a distributed process should be able to do so through a simple, uniform interface (declare what messages to receive and send messages to named peers) without knowing anything about how that delivery is performed. The transport responsibility (peer discovery, routing, serialization) is entirely encapsulated in a dedicated gateway node that runs alongside the existing drone stack.

The second is \textbf{non-invasive integration}. The gateway extends system capability without modifying existing components. A system that does not require inter-agent communication is unaffected; components that do opt in through a thin client library without disrupting existing functionality. This constitutes a brownfield integration strategy: distributed-process capability is introduced incrementally through new behaviors, while all pre-existing system functionality remains unmodified.

\subsection{Architecture}

\begin{figure}[t]
\centering
\begin{tikzpicture}[
  font=\scriptsize,
  box/.style={draw, fill=white, rectangle, minimum width=1.4cm,
              minimum height=0.45cm, align=center, inner sep=2pt},
  ll/.style={densely dashed, black!40},
  arr/.style={-{Stealth[length=3pt,width=2.5pt]}, semithick},
  marr/.style={-{Stealth[length=3pt,width=2.5pt]}, semithick, dashed},
  narr/.style={-{Stealth[length=3pt,width=2.5pt]}, semithick, dashed, black!55},
  lbl/.style={font=\tiny},
  sep/.style={dotted, black!50, semithick},
  phase/.style={font=\tiny\itshape, black!75},
]
  \def\xA{0}    
  \def\xB{1.9}  
  \def\xC{3.7}  
  \def\xD{5.5}  

  \node[box] at (\xA,0.2) (pA) {Behavior};
  \node[box] at (\xB,0.2) (pB) {SL Client};
  \node[box] at (\xC,0.2) (pC) {SL Node};
  \node[box] at (\xD,0.2) (pD) {\shortstack{Remote\\[-1pt]Agent}};

  \foreach \x in {\xA,\xB,\xC,\xD}
    \draw[ll] (\x,-0.1) -- (\x,-6.45);

  \draw[rounded corners=3pt, black!25]
    (-0.95,1.0) rectangle (4.65,-6.62);
  \node[font=\tiny\bfseries, text=black!70, anchor=north]
    at ({(\xA+\xC)/2}, 1.0) {Drone A};
  \draw[rounded corners=3pt, black!25]
    (4.65,1.0) rectangle (6.50,-6.62);
  \node[font=\tiny\bfseries, text=black!70, anchor=north]
    at (\xD, 1.0) {Peer};

  \draw[sep] (-0.95,-0.38) -- (6.35,-0.38);
  \node[phase, anchor=east] at (-0.95,-0.64) {(1) Register};

  \draw[arr] (\xA,-0.72) -- (\xB,-0.72)
    node[lbl, above, midway] {\texttt{register(T,\,cb)}};
  \draw[marr] (\xB,-1.22) -- (\xC,-1.22)
    node[lbl, above, midway] {\texttt{subscribe(T)}};

  \draw[sep] (-0.95,-1.62) -- (6.35,-1.62);
  \node[phase, anchor=east, align=right] at (-0.95,-1.88) {(2) Receive\\ message};

  \draw[narr] (\xD,-2.10) -- (\xC,-2.10)
    node[lbl, above, midway, text=black] {msg of type \texttt{T}};
  \draw[marr]  (\xC,-2.60) -- (\xB,-2.60)
    node[lbl, above, midway] {dispatch(msg)};
  \draw[arr]  (\xB,-3.10) -- (\xA,-3.10)
    node[lbl, above, midway] {\texttt{cb(msg)}};

  \draw[sep] (-0.95,-3.55) -- (6.35,-3.55);
  \node[phase, anchor=east] at (-0.95,-3.80) {(3) Forward};

  \draw[arr] (\xA,-4.00) -- (\xB,-4.00)
    node[lbl, above, midway] {\texttt{forward(T,\,msg)}};
  \draw[marr] (\xB,-4.55) -- (\xC,-4.55)
    node[lbl, above, midway] {route(msg)};
  \draw[narr] (\xC,-5.10) -- (\xD,-5.10)
    node[lbl, above, midway, text=black] {[\,network\,]};

  \fill[black!12] (\xD-0.10,-5.30) rectangle (\xD+0.10,-5.80);
  \draw[black!55] (\xD-0.10,-5.30) rectangle (\xD+0.10,-5.80);
  \node[lbl, right] at (\xD+0.13,-5.55) {\texttt{cb(msg)}};

  \draw[arr]  (0.15,-6.15) -- (0.75,-6.15);
  \node[lbl, anchor=west] at (0.80,-6.15) {method call};
  \draw[marr] (0.15,-6.45) -- (0.75,-6.45);
  \node[lbl, anchor=west] at (0.80,-6.45) {message};

\end{tikzpicture}
\caption{SwarmLink message flow. Solid arrows denote method invocations (Behavior\,--\,SL\,Client); dashed arrows denote message exchanges (SL\,Client\,--\,SL\,Node\,--\,Remote\,Agent). \textit{(1)~Register}: a behavior registers callback \texttt{cb} for message type \texttt{T}; the SwarmLink Client subscribes at the node level. \textit{(2)~Receive~message}: when a peer sends a message of type \texttt{T}, the SwarmLink Node dispatches it to the client, which invokes \texttt{cb} directly. \textit{(3)~Forward}: a behavior calls \texttt{forward}; SwarmLink serializes and routes the message to all reachable peers. In all cases the behavior has no knowledge of peer addresses or transport.}
\label{fig:gateway}
\end{figure}
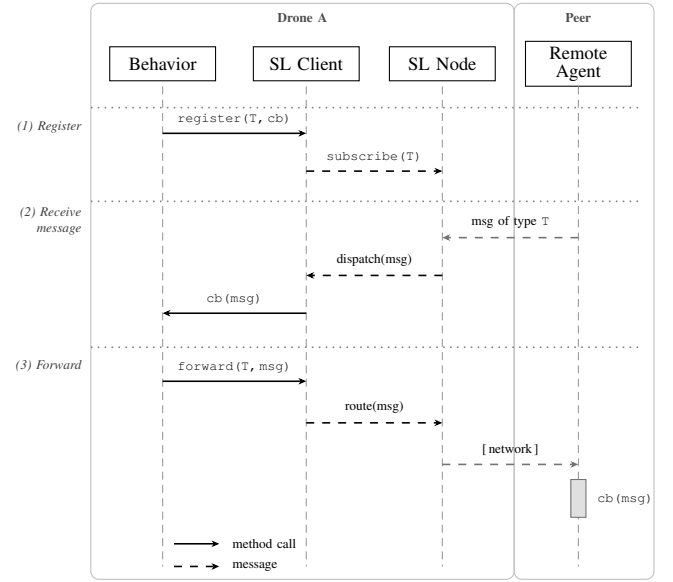

The SwarmLink comprises two artifacts  interacting as shown in Fig.~\ref{fig:gateway}:

\textbf{SwarmLink Node}: one instance runs per drone and owns all inter-agent traffic. It continuously scans the active network, creates forwarding channels to newly discovered peers, and prunes channels for peers that have disappeared. This scan provides both runtime peer discovery and implicit failure detection: a crashed or disconnected drone ceases to be reachable within one discovery cycle, and its channel is automatically removed.

\textbf{SwarmLink Client}: a lightweight library that any behavior embeds to participate in distributed processes. It exposes two operations: \textit{register} declares interest in a message type and provides a callback invoked on arrival; \textit{forward} sends a typed message to a named set of peers. The client handles serialization and delegates delivery to the local SwarmLink Node. From the behavior's perspective, peer addresses, network topology, and transport details are completely hidden.

The SwarmLink is framework-agnostic: any system that can host a companion node and link a client library can adopt it. Section~\ref{sec:auction} shows what this enables when applied to Aerostack2.

Compared to direct use of ROS~2 cross-namespace topics, SwarmLink operates at a higher level of abstraction. Raw DDS communication requires per-message-type topic definitions, static publisher-subscriber wiring, and explicit reconnection upon fleet reconfiguration; consulting the ROS~2 graph at runtime to discover active nodes also incurs repeated query overhead that grows with fleet size. The SwarmLink Client abstracts this to two fleet-size-agnostic calls, \textit{register} and \textit{forward}, applicable uniformly across coordination protocols without per-protocol wiring; behavior code is identical in simulation (shared machine) and field deployment (Wi-Fi). In the Aerostack2 integration, the SwarmLink Node uses ROS~2 DDS topics as the transport backend; the transport-agnostic interface permits future substitution with alternatives such as Zenoh or MQTT without modifying behavior code. Table~\ref{tab:swarmlink_vs_dds} summarises the key differences.

\begin{table}[t]
\caption{SwarmLink vs.\ direct ROS~2 DDS usage.}
\label{tab:swarmlink_vs_dds}
\begin{center}
\small
\renewcommand{\arraystretch}{1.2}
\begin{tabular}{p{2.0cm}p{2.4cm}p{2.4cm}}
\hline
\textbf{Dimension} & \textbf{SwarmLink} & \textbf{Raw ROS~2 DDS} \\
\hline
Discovery overhead & Registry resolved once at send time & Graph queried per message type at runtime \\
Naming coupling & Typed registry; no topic names in behavior code & Per-message-type topic names hard-coded \\
Transport flexibility & Backend swappable (DDS, Zenoh, MQTT) & Tied to DDS \\
Fleet reconfiguration & Handled automatically by SwarmLink Node & Explicit publisher/subscriber rewiring required \\
\hline
\end{tabular}
\end{center}
\end{table}

\begin{figure}[t]
\centering
\includegraphics[width=\columnwidth]{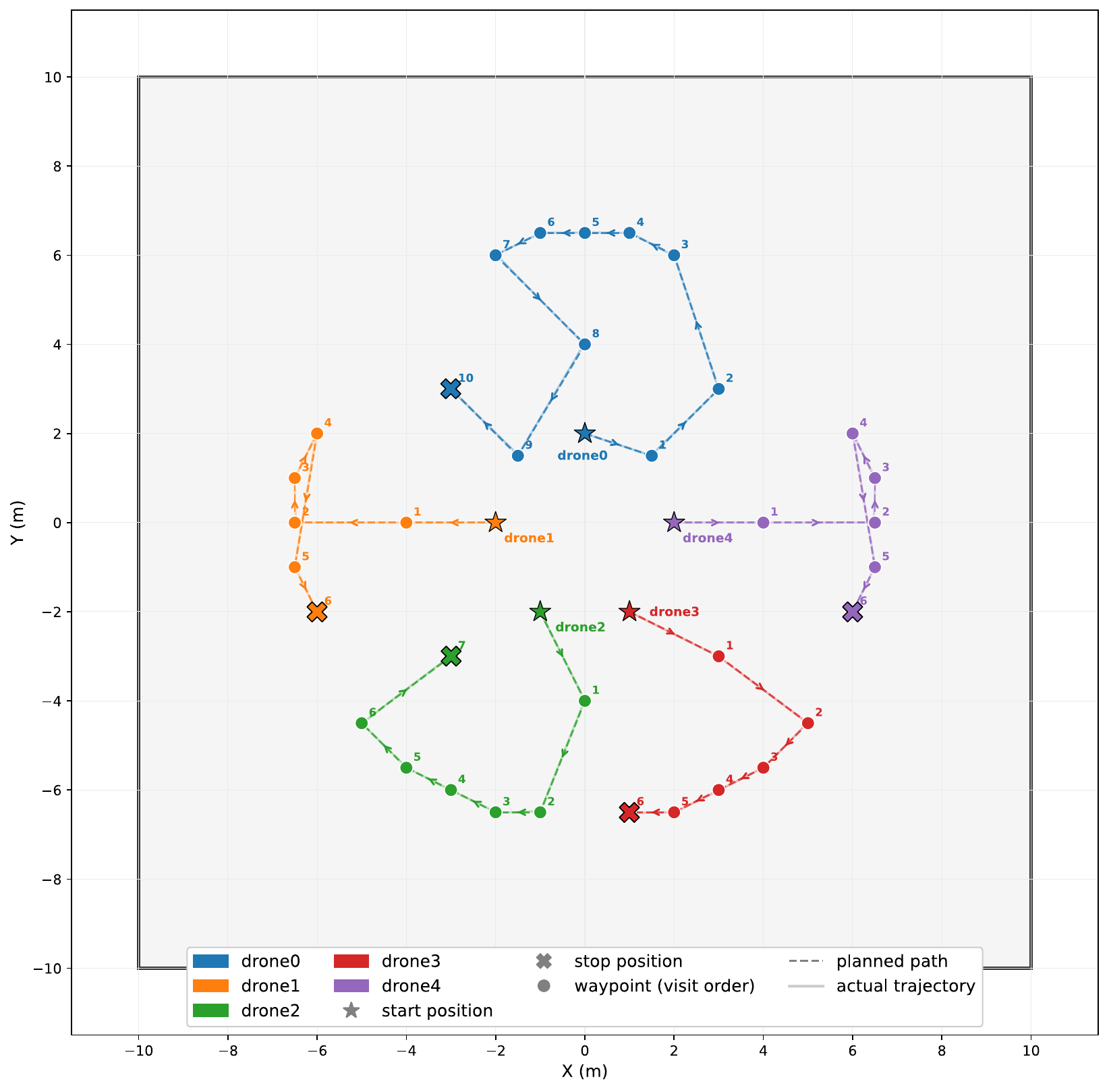}
\caption{Sequential greedy auction with five drones ($N=5$, $M=30$ waypoints). Each color corresponds to one drone's assigned route; waypoints are visited in the order determined by the auction. The partition minimizes total travel distance greedily, with each drone claiming the nearest unclaimed waypoint in each round.}
\label{fig:penta}
\end{figure}

\section{Distributed Task Assignment: The Auction Behavior}
\label{sec:auction}

The second contribution of this paper is a non-invasive extension of Aerostack2 \cite{fernandezcortizas2024aerostack2softwareframeworkdeveloping} that addresses the runtime problems identified in the Introduction, without any modification to the framework core. The SwarmLink Node is deployed as a companion process alongside each drone's Aerostack2 stack; the SwarmLink Client is linked into new Aerostack2 behaviors. All pre-existing single-drone and choreographed multi-drone workflows continue to function identically. On top of this unchanged foundation, the Auction Behavior implements distributed task assignment: partitioning inspection panels among the active fleet, recovering from drone failures, and responding to image-quality events. In every case the task assignment problem reduces to assigning a set of uninspected waypoints to a set of active drones, and the SwarmLink is what makes a fully distributed solution possible.

\subsection{Problem and Algorithm}

Given $N$ active UAVs and a set $\mathcal{W} = \{w_1, \ldots, w_M\}$ of uninspected panel-row waypoints, we seek a partition $\mathcal{P} = \{P_1, \ldots, P_N\}$ with $\bigcup_i P_i = \mathcal{W}$ that minimizes total travel cost. We use a sequential greedy auction \cite{1677943,Otte2020}: each round every agent bids for its preferred unclaimed waypoint, the one nearest to its current position, using Euclidean distance as the cost metric, the winner broadcasts its claim, and rounds continue until all waypoints are assigned. Message complexity is $O(N \cdot M)$, well suited to the fleet sizes and panel counts of PV inspection scenarios. A centralized planner can compute an optimal partition but constitutes a single point of failure: if the coordinator is lost the entire mission stalls. The sequential greedy auction does not guarantee optimal partitions \cite{1677943}; it yields a feasible allocation that degrades gracefully under agent failures --- surviving agents converge without any central coordinator --- and its simplicity and low message complexity make it well suited to the real-time constraints of in-mission recovery. The plugin architecture allows substitution with algorithms offering stronger guarantees, such as CBBA \cite{5072249}, without modifying SwarmLink or the rest of the system. Fig.~\ref{fig:penta} shows the algorithm partitioning 30 waypoints among five drones in simulation.

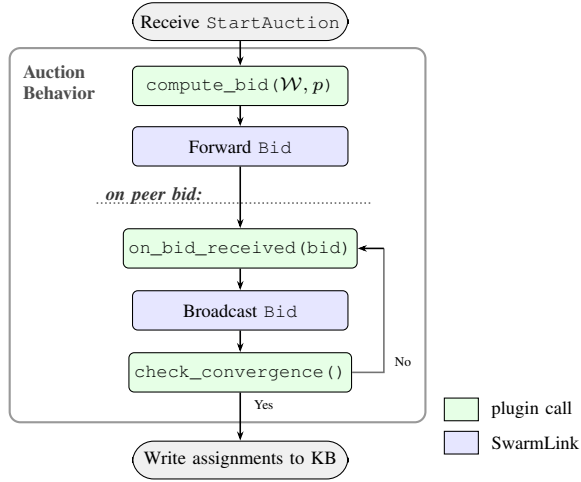
\begin{figure}[tbh]
\centering
\begin{tikzpicture}[
  font=\scriptsize,
  sstep/.style={draw, rounded corners=2pt, fill=blue!10,
                minimum width=2.85cm, minimum height=0.52cm,
                align=center, inner sep=2pt},
  pstep/.style={draw, rounded corners=2pt, fill=green!10,
                minimum width=2.85cm, minimum height=0.52cm,
                align=center, inner sep=2pt},
  term/.style={draw, rounded corners=8pt, fill=black!6,
               minimum width=2.85cm, minimum height=0.50cm,
               align=center, inner sep=2pt},
  arr/.style={-{Stealth[length=3.5pt,width=2.5pt]}, semithick},
  back/.style={semithick, black!55},
]
  \def\fc{1.55}   

  \draw[rounded corners=4pt, black!40, thick]
    (-1.5, -0.35) rectangle (4, -5.3);
  \node[font=\scriptsize\bfseries, text=black!70, fill=white, inner sep=2pt,
        align=left, anchor=west] at (-1.4, -0.8) {Auction\\Behavior};

  \node[term]  (start) at (\fc,  0.00) {Receive \texttt{StartAuction}};
  \node[pstep] (cb)    at (\fc, -0.85)
    {\texttt{compute\_bid(}$\mathcal{W},p$\texttt{)}};
  \node[sstep] (fwd1)  at (\fc, -1.65)
    {Forward \texttt{Bid}};

  \draw[black!60, semithick, densely dotted] (-0.35,-2.4) -- (3.30,-2.4);
  \node[font=\scriptsize\bfseries\itshape, black!80, anchor=west] at (-0.35,-2.3)
    {on peer bid:};

  \node[pstep] (obr)  at (\fc, -3.00)
    {\texttt{on\_bid\_received(bid)}};
  \node[sstep] (brd)  at (\fc, -3.82)
    {Broadcast \texttt{Bid}};
  \node[pstep] (cc)   at (\fc, -4.64)
    {\texttt{check\_convergence()}};
  \node[term]  (done) at (\fc, -5.80)
    {Write assignments to KB};

  \draw[arr] (start.south) -- (cb.north);
  \draw[arr] (cb.south)    -- (fwd1.north);
  \draw[arr] (fwd1.south)  -- (obr.north);
  \draw[arr] (obr.south)   -- (brd.north);
  \draw[arr] (brd.south)   -- (cc.north);
  \draw[arr] (cc.south)    -- (done.north);
  \node[font=\tiny, anchor=west] at (\fc+0.06,-5.07) {Yes};

  \draw[back] (cc.east)  -- (3.45,-4.64) -- (3.45,-3.00);
  \draw[arr]  (3.45,-3.00) -- (obr.east);
  \node[font=\tiny, anchor=west] at (3.47,-4.50) {No};

  \fill[green!10] (4.25,-5.28) rectangle (4.70,-4.98);
  \draw           (4.25,-5.28) rectangle (4.70,-4.98);
  \node[font=\scriptsize, anchor=west] at (4.75,-5.13) {plugin call};

  \fill[blue!10]  (4.25,-5.72) rectangle (4.70,-5.42);
  \draw           (4.25,-5.72) rectangle (4.70,-5.42);
  \node[font=\scriptsize, anchor=west] at (4.75,-5.57) {SwarmLink};

\end{tikzpicture}
\caption{Auction Behavior execution loop. On \texttt{StartAuction}, each drone calls \texttt{compute\_bid} to select its preferred waypoint and forwards the result via SwarmLink. Whenever a peer's \texttt{Bid} arrives, \texttt{on\_bid\_received} is called and the updated bid is rebroadcast; \texttt{check\_convergence} then tests whether all waypoints are claimed. The loop repeats until every drone converges, after which assignments are written to the knowledge base. Green steps are delegated to the \texttt{AuctionPlugin} interface, making the bidding strategy independently swappable.}
\label{fig:auction}
\end{figure}

\subsection{Auction Protocol via SwarmLink}
\begin{figure*}[t!]
\centering
\subfloat[Initial allocation: five drones cover all waypoints.\label{fig:prereplan}]{%
  \includegraphics[width=0.45\textwidth]{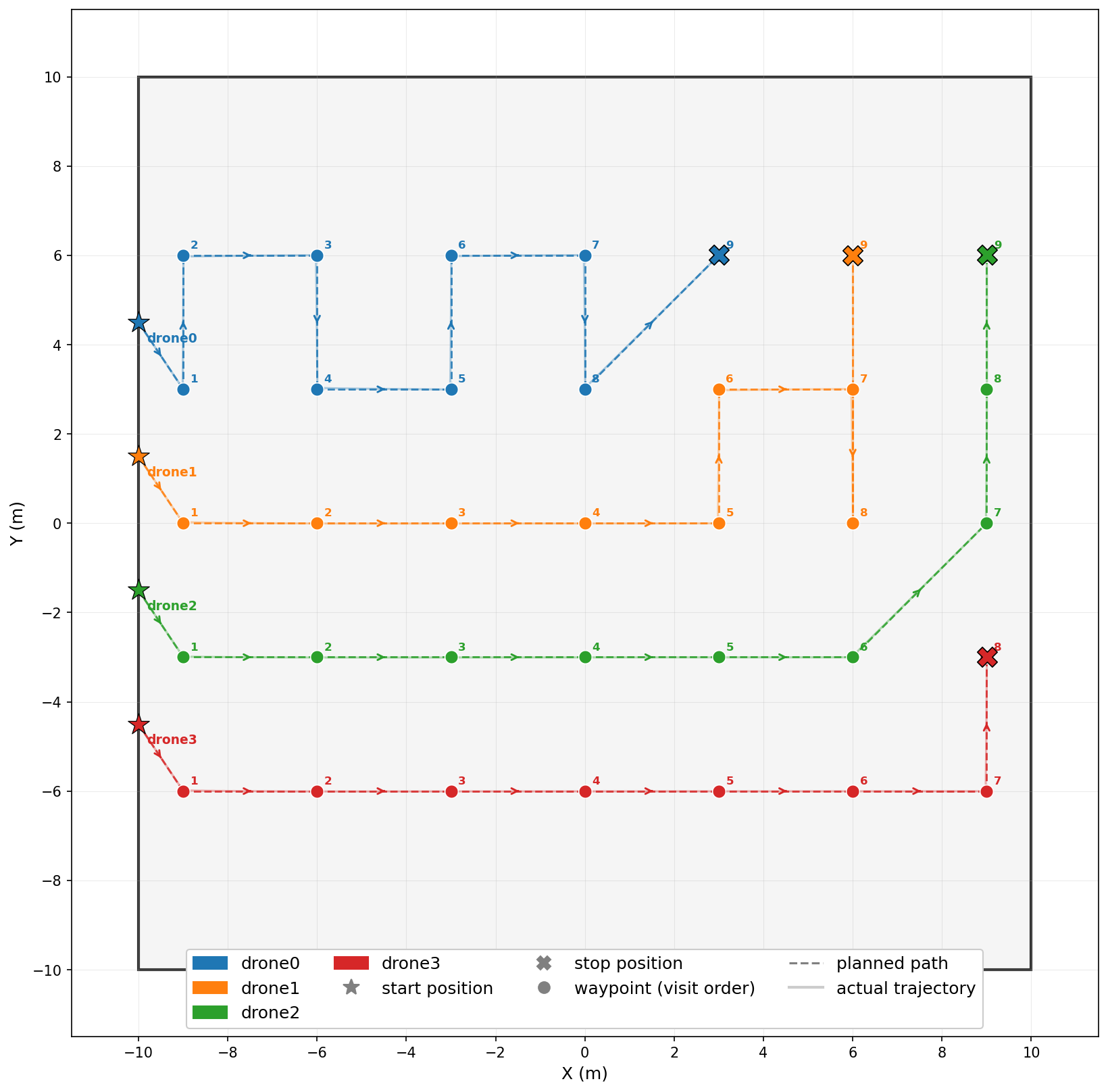}}
\hfill
\subfloat[After drone1 failure: surviving drones re-auction the remaining panels.\label{fig:replan}]{%
  \includegraphics[width=0.45\textwidth]{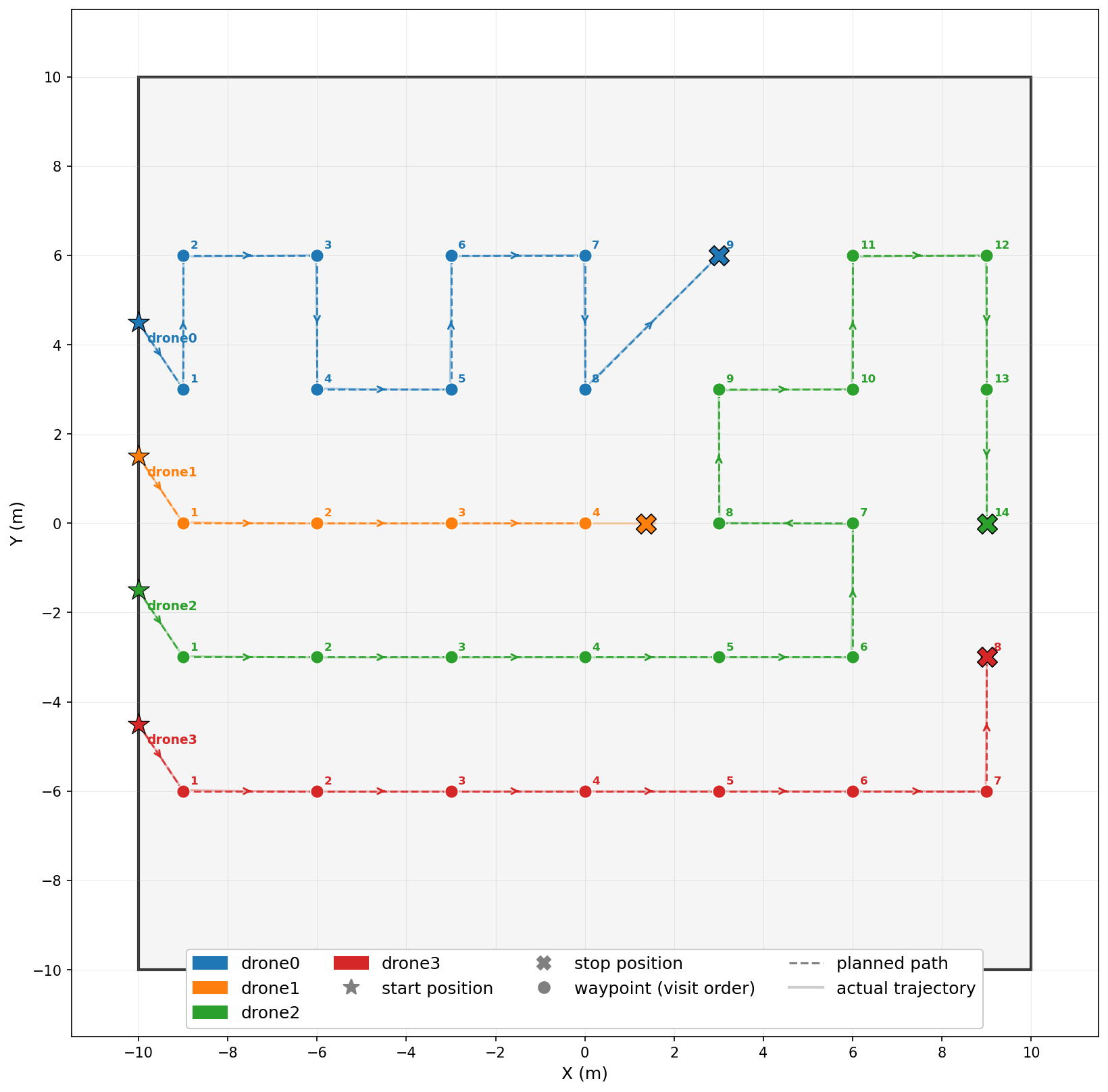}}
\caption{Distributed failure recovery via SwarmLink. \textit{Symbol key}: dotted lines --- auction-assigned planned order; solid lines --- actual flight trajectories; \texttimes{} --- failed drone. \textit{(a)}~Four drones execute their auction-assigned routes. \textit{(b)}~Drone1 fails mid-mission; SwarmLink detects the loss within one discovery cycle and the surviving drones re-auction drone1's unfinished waypoints, resuming inspection without operator intervention. The image-quality failure scenario follows the same sequence: the affected drone's panels are removed from the knowledge base and re-auctioned among the active fleet. }
\label{fig:failure_recovery}
\end{figure*}

The auction involves two message types: a \texttt{StartAuction} broadcast sent once by the auctioneer to initiate bidding, and \texttt{Bid} messages broadcast by each agent to claim a waypoint. Without an inter-agent abstraction layer, each participant must maintain explicit connections to all peers, re-establish them upon fleet reconfiguration, and replicate the connection logic for each coordination protocol, yielding an $O(N^2)$ wiring burden incompatible with dynamic fleet management. Fig.~\ref{fig:auction} shows the Auction Behavior architecture, with bidding strategy and cost evaluation encapsulated in independently swappable plugins.

SwarmLink reduces each step to a single call. The auctioneer calls \textit{forward}(\texttt{StartAuction}, \textit{peers}) once; SwarmLink resolves the current set of reachable peers at runtime and delivers the message to each without the auctioneer knowing their addresses. Each bidder calls \textit{register}(\texttt{Bid}, \textit{callback}) once; its callback fires whenever any peer broadcasts a bid, regardless of fleet size or composition.

Bid propagation works identically: on claiming a waypoint an agent calls \textit{forward}(\texttt{Bid}, \textit{all\_peers}), and SwarmLink delivers it to every active participant. Because SwarmLink continuously tracks peer reachability, the protocol is inherently failure-aware: a drone that crashes mid-auction is pruned from the active peer set within one discovery cycle, and its bids simply stop arriving. The remaining agents converge without any explicit failure-handling logic.

The one protocol-level constraint introduced by this communication model is ordering: since the auctioneer sends \texttt{StartAuction} and its own first \texttt{Bid} back-to-back, each bidder must fully initialize its waypoint list \textit{within} the \texttt{StartAuction} callback before the bid loop starts. Deferring this initialization would cause the first bid to arrive before the bidder is ready, deadlocking the auction. SwarmLink's delivery ordering guarantee makes this manageable: both messages arrive in send order, so synchronous initialization inside the callback is sufficient.

On convergence, each agent writes its assignments to the shared knowledge base, a per-drone store that holds the current agreed mission state from which downstream path-planning behaviors retrieve their flight plan. The same SwarmLink registrations remain active and are immediately reusable if a failure or image-quality event triggers a re-auction.

\textbf{Auction initiation.} Any drone that detects a peer failure or an image-quality event initiates an auction; there is no permanent auctioneer. To handle the case where two drones detect the same event simultaneously and both emit a \texttt{StartAuction} message, each message carries a unique identifier composed of the initiator's ID and a timestamp. A drone that receives a \texttt{StartAuction} while already running an auction compares identifiers and defers to the one with the lower initiator ID, collapsing both triggers into a single auction run. No leader-election round is required; the lightweight deduplication is sufficient for the fleet sizes and failure rates of PV inspection missions.

\subsection{Handling Failures and Image Quality Events}

\textbf{Drone failure.} The SwarmLink Node detects a failed peer within one discovery cycle and prunes its channel. The surviving drones identify the unfinished waypoints belonging to the lost agent in the knowledge base and launch a new Auction Behavior instance with that subset and the surviving fleet. The re-auction is structurally identical to the initial one. Fig.~\ref{fig:failure_recovery} illustrates this recovery in simulation: drone1 fails mid-mission and the two surviving drones re-auction the abandoned panels without ground-station involvement.

\textbf{Image quality failure.} When an onboard image-quality evaluation behavior classifies captured images as invalid \cite{JaviCPP}, the drone removes the affected panels from the knowledge base and broadcasts a reassignment request via the SwarmLink Client. The remaining drones launch a new Auction Behavior for those panels, ensuring full inspection coverage without operator intervention.

The plugin architecture ensures that alternative assignment strategies, such as the Consensus-Based Bundle Algorithm \cite{5072249} or energy-aware cost functions, can be substituted without modifying the SwarmLink or the rest of the system, consistent with the non-invasive design principle of the first contribution. Together, the two failure-recovery mechanisms confirm the second contribution: Aerostack2 has gained all three desired inspection properties, ground-station independence, drone-failure resilience, and image-quality assurance, purely through the addition of the SwarmLink and the Auction Behavior, with zero changes to the existing framework.

\section{Conclusion}

This paper presented SwarmLink and its application to distributed multi-UAV photovoltaic inspection. SwarmLink is a non-invasive inter-agent communication infrastructure that allows any component of an existing aerial framework to participate in distributed coordination through a uniform \textit{register}/\textit{forward} interface, without knowledge of the underlying transport and without any modification to the host system. Applied to Aerostack2, it enables the Auction Behavior: a distributed sequential greedy auction that unifies sensing-target allocation, platform-failure recovery, and data-quality-triggered reassignment into a single runtime re-auction mechanism, requiring zero changes to the Aerostack2 core and no ground-station involvement during the mission.

Three limitations are acknowledged. First, the sequential greedy auction does not guarantee optimal partitions; the plugin architecture supports substitution with algorithms offering stronger guarantees, such as CBBA \cite{5072249}. Second, the protocol assigns the auctioneer role to a single drone; auctioneer failure mid-round is not currently handled and is a planned extension. Third, a network partition that splits the swarm into two disconnected subsets will cause each subset to independently re-auction all remaining waypoints, leading to duplicated coverage and potential conflicts upon reconnection. Mitigations under investigation include heartbeat-based partition detection, goal-ID deduplication upon merge, and a lightweight consensus round when subsets rejoin; resolving this is a known open problem in fully decentralised coordination \cite{Otte2020} and is left for future work.

Future work includes field validation with a physical RPAS fleet, evaluation of alternative auction strategies, and scalability analysis as fleet size and panel count grow.

\section*{Acknowledgment}

This work is supported by the CORESENSE project funded by the European Union under Horizon Europe grant agreement No.~101070254 (HORIZON-CL4-2021-DIGITAL-EMERGING-01-11). The authors thank the CORESENSE consortium for their contributions to the inspection testbed specification this work is based on. The work of the first author is supported by the Community of Madrid under its Program for Predoctoral Researcher with Ref. PIPF-2023/TEC-31167 (BOCM - 3454/2024)

\bibliographystyle{IEEEtran}
\bibliography{biblio}

\end{document}